\documentclass[letterpaper, 10 pt, conference]{ieeeconf}  

\IEEEoverridecommandlockouts                              

\usepackage{times} 
\usepackage{amsmath} 
\usepackage{amssymb}  

\usepackage[subrefformat=parens]{subcaption}
\usepackage{balance}
\usepackage{cite}
\usepackage{graphicx}
\usepackage{booktabs}
\usepackage{tabularx}
\usepackage{multirow}
\usepackage{multicol}
\usepackage{cite}
\usepackage{bm}
\usepackage{url}
\usepackage{balance}
\usepackage{color}
\usepackage{bbding}
\usepackage{nccmath}
\usepackage{xspace}
\usepackage{wrapfig}
\usepackage{colortbl}
\usepackage[capitalize]{cleveref}
\crefname{section}{Sec.}{Secs.}
\Crefname{section}{Section}{Sections}
\Crefname{table}{Table}{Tables}
\crefname{table}{Tab.}{Tabs.}

\xspaceremoveexception{-}

\newcommand{\methodname}{ViewBirdiformer\xspace}
\newcommand{\motionmodelname}{TransMotion\xspace}
\newcommand{\baselinename}{GeoVB}

\def\eg{{\it e.g.,~}}

\def\ie{{\it i.e.,~}}
\def\etal{{\it et al.~}}

\def\Pose{\Pi}
\def\pose{\bm \pi}

\def\traj{\mathcal X}

\def\States{\mathcal {\bm S}}
\def\state{\bm s}
\def\ped{\bm x}
\def\pedi{\bm p} 
\def\pedc{i} 

\def\Rerr{\Delta \bm r}
\def\Terr{\Delta \bm t}
\def\Xerr{\Delta \bm x}
\def\XerrRel{\Delta \bm {\tilde x}}
\def\Xgt{\dot{\bm x}}

\title{\LARGE \bf ViewBirdiformer: Learning to recover ground-plane crowd trajectories and ego-motion from a single ego-centric view}

\author{Mai Nishimura$^{1,2}$, Shohei Nobuhara${^2}$ and Ko Nishino$^{2}$
\thanks{$^{1}$Mai Nishimura is with OMRON SINIC X Corporation, 5-24-5, Hongo, Bunkyo-ku, Tokyo, Japan
        {\tt\small mai.nishimura@sinicx.com}}%
\thanks{$^{2}$Shohei Nobuhara and Ko Nishino are with Kyoto University, Yoshida Honmachi, Sakyo-ku, Kyoto, Japan
        {\tt\small \{nob,kon\}@i.kyoto-u.ac.jp}}%
}

\begin{document}

\maketitle
\thispagestyle{empty}
\pagestyle{empty}

\begin{abstract}
We introduce a novel learning-based method for view birdification~\cite{nishimura2021bmvc}, the task of recovering ground-plane trajectories of pedestrians of a crowd and their observer in the same crowd just from the observed ego-centric video. View birdification becomes essential for mobile robot navigation and localization in dense crowds where the static background is hard to see and reliably track. It is challenging mainly for two reasons; i) absolute trajectories of pedestrians are entangled with the movement of the observer which needs to be decoupled from their observed relative movements in the ego-centric video, and ii) a crowd motion model describing the pedestrian movement interactions is specific to the scene yet unknown a priori. 
For this, we introduce a Transformer-based network referred to as \methodname which implicitly models the crowd motion through self-attention and decomposes relative 2D movement observations onto the ground-plane trajectories of the crowd and the camera through cross-attention between views.
Most important, \methodname achieves view birdification in a single forward pass which opens the door to accurate real-time, always-on situational awareness. 
Extensive experimental results demonstrate that \methodname achieves accuracy similar to or better than state-of-the-art with three orders of magnitude reduction in execution time.
\end{abstract}

\section{Introduction}
\label{sec:intro}
\begin{figure*}[t]
    \centering
   \includegraphics[width=0.8\linewidth]{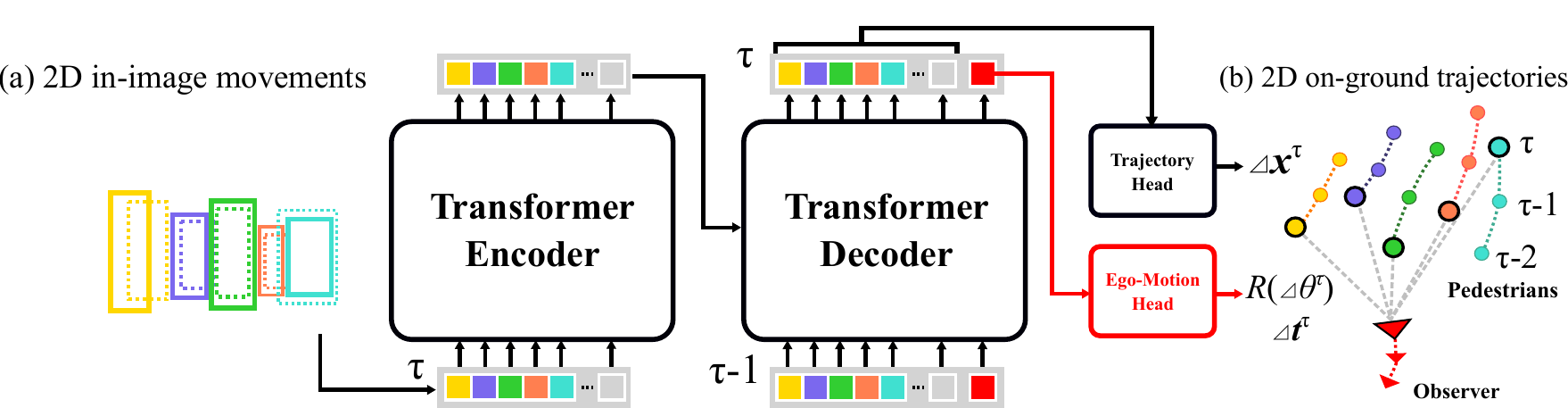}
    \caption{Given bounding boxes of moving pedestrians in an ego-centric view captured in the crowd, \methodname reconstructs on-ground trajectories of both the observer and the surrounding pedestrians.}
    \label{fig:teaser}
\end{figure*}

\looseness=-1
We as human beings have a fairly accurate idea of the absolute movements of our surroundings in the world coordinate frame, even when we can only observe their movements relative to our own in our sight such as when walking in a crowd.
Enabling a mobile agent to maintain a dynamically updated map of surrounding absolute movements on the ground, solely from observations collected from its own vantage point, would be of significant use for various applications including robot navigation~\cite{nishimura2020iros}, autonomous driving~\cite{lee2020pillarflow}, sports analysis~\cite{cioppa2021camera}, and crowd monitoring~\cite{gupta2018social,ivanovic2019trajectron,mehran2009abnormal}. 
The key challenge lies in the fact that when the observer (\eg person or robot) is surrounded by other dynamic agents, static ``background'' can hardly be found in the agent's field of view. In such scenes, conventional visual localization methods including SLAM would fail since static landmarks become untrackable due to frequent occlusions by pedestrians and the limited dynamically changing field of view~\cite{nishimura2021bmvc}.
External odometry signals such as IMU and GPS are also often unreliable. Even when they are available, visual feedback becomes essential for robust pose estimation (imagine walking in a crowd with closed eyes).

\looseness=-1
Nishimura \etal recently introduced this exact task as \emph{view birdification} whose goal is to recover on-ground trajectories of a camera and a crowd just from perceived movements (not appearance) in an ego-centric video~\cite{nishimura2021bmvc}
\footnote{Note that Bird's Eye View transform is a completely different problem as it concerns a single frame view of the appearance (not the movements) and cannot reconstruct the camera ego-motion.}.
They proposed to decompose these two types of trajectories, one of the pedestrians in the crowd and another of a person or mobile robot with an ego-view camera, with a cascaded optimization which alternates between estimating the displacements of the camera and estimating those of surrounding pedestrians while constraining the crowd trajectories with a pre-determined crowd motion model~\cite{helbing1995social,scholler2020constant}. 
This iterative approach suffers from two critical problems which hinder their practical use. First, its iterative optimization incurs a large computational cost which precludes real-time use. Second, the analytical crowd model as a prior is restricting and not applicable to diverse scenes where the crowd motion model is unknown.

In this paper, we propose \emph{ViewBirdiformer}, a Transformer-based view birdification method.
Instead of relying on restrictive assumptions on the motion of surrounding people and costly alternating optimization, we define a Transformer-based network that learns to reconstruct on-ground trajectories of the surrounding pedestrians and the camera from a single ego-centric video while simultaneously learning their motion models.
As \cref{fig:teaser} depicts, \methodname takes in-image 2D pedestrian movements as inputs, and outputs 2D pedestrian trajectories and the observer's ego-motion on the ground plane. The multi-head self-attention on the motion feature embeddings of each pedestrian of \methodname captures the local and global interactions of pedestrians. At the same time, it learns to reconstruct on-ground trajectories from observed 2D motion in the image with cross-attention on features coming from different viewpoints.

A key challenge of this data-driven view birdification lies in the inconsistency of coordinate frames between input and output movements---the input is 2D in-image movements relative to ego-motion, but the expected outputs are on-ground trajectories in absolute coordinates (\ie independent of the observer's motion). 
\methodname resolves this by introducing the two types of queries, \ie the camera ego-motion and pedestrian trajectories, in a multi-task learning formulation, and by transforming coordinates of pedestrian queries relative to the previous ego-motion estimates.

\looseness=-1
We thoroughly evaluate the effectiveness of our method using the view birdification dataset~\cite{nishimura2021bmvc} and also by conducting ablation studies which validate its key components.
The proposed Transformer-based architecture learns to reconstruct trajectories of the camera and the crowd while learning their motion models by adaptively attending to movement features of them in the image plane and on the ground. 
It enables real-time view birdification of arbitrary ego-view crowd sequences in a single inference pass, 
which leads to three orders of magnitude speedup from the iterative optimization approach~\cite{nishimura2021bmvc}.
We show that the results of \methodname can be opportunistically refined with geometric post-processing, which results in similar or better accuracy than state-of-the-art~\cite{nishimura2021bmvc} but still in orders of magnitude faster execution time.

\section{Related Work}

\looseness=-1
\subsection{View Birdification}
As summarized in \Cref{tab:relevant-tasks}, View Birdification~\cite{nishimura2021bmvc} is not the same as bird's-eye view (BEV)
transformation~\cite{yang2021projecting,fiery2021,zhou2022cross,saha2022translating}. 
BEV transformation refers to the task of rendering a 2D top-down view image from an on-ground ego-centric view and concerns the appearance of the surroundings as seen from the top
and does not resolve the ego-motion, \ie all recovered BEVs are still relative to the observer. 
View birdification, in contrast, reconstructs both the observer's and surrounding pedestrians' locations on the ground so that the relative movements captured in the ego-centric view can be analyzed in a single world coordinate frame on the ground (\ie ``birdified''). View birdification thus fundamentally differs from BEV transform as it is inherently a 3D transform that accounts for the ego-motion, \ie the 2D projections of surrounding people in the 2D ego-view need to be implicitly or explicitly lifted into 3D and translated to cancel out the jointly estimated ego-motion of the observer before being projected down onto the ground-plane. Nishimura \etal introduced a geometric method for view birdification~\cite{nishimura2021bmvc}, which explicitly transforms the 2D projected pedestrian movements into 3D but on the ground plane with a graph energy minimization by leveraging analytically expressible crowd motion models~\cite{helbing1995social}. Our method fundamentally differs from this in that the transformation from 2D in-image movement to on-ground motion as well as the on-ground coordination of pedestrian motion is jointly learned from data.

\begin{table}[t]
  \centering
    \caption{
  View birdification (VB) is the only task that simultaneously recovers the absolute trajectories of the camera and its surrounding pedestrians only from their perceived movements relative to an observer. 
  }
    \label{tab:relevant-tasks}
  \scriptsize
  \newcolumntype{Y}{>{\centering\arraybackslash}X}
  \begin{tabularx}{\linewidth}{l|cc|cc|cY}
    \toprule[1pt]
    \textbf{Task} &\multicolumn{2}{c|}{\textbf{Input}} & \multicolumn{2}{c|}{\textbf{Output}} & \multicolumn{2}{c}{\textbf{Scenes}} \\
    & static & dynamic & ego & traj & a few people & crowd \\
    \midrule
    BEV~\cite{yang2021projecting} & \checkmark & & & & \checkmark \\ 
    3D MOT~\cite{Hu2021QD3DT} & & \checkmark &  & \checkmark &\checkmark & \checkmark \\
    SLAM~\cite{huang2020clustervo} & \checkmark & \checkmark& \checkmark & \checkmark & \checkmark &  \\
   \rowcolor[rgb]{0.93,1.0,0.87} VB~(\textbf{Ours},\cite{nishimura2021bmvc}) &  & \checkmark & \checkmark & \checkmark & \checkmark & \checkmark \\
    \bottomrule
  \end{tabularx}
\end{table}

\subsection{Simultaneous Localization and Mapping (SLAM)} 
Dynamic SLAM and its variants inherently rely on the assumption that the world is static~\cite{murorb2,engel2014lsd,karkus2021differentiable}. Dynamic objects cause feature points to drift and contaminate the ego-motion estimate and consequently the 3D reconstruction. Past methods have made SLAM applicable to dynamic scenes, ``despite'' these dynamic objects, by treating them as outliers~\cite{hahnel2003map} or explicitly tracking and filtering them ~\cite{bescos2018dynaslam,yu2018ds,vincent2020dynamic,dot2021icra}. A notable exception is Dynamic Object SLAM which explicitly incorporates such objects into its geometric optimization~\cite{huang2020clustervo,yang2019cubeslam,henein2020dynamic}. The method detects and tracks dynamic objects together with static keypoints, but assumes that the dynamic objects in view are rigid and obey a simple motion model that results in smoothly changing poses. 
None of the above methods consider the complex pedestrian interactions in the crowd~\cite{helbing1995social,van2011reciprocal,gupta2018social,yuan2021agent}.
Our method fundamentally differs from dynamic SLAM in that it reconstructs both the observer's ego-motion and the on-ground trajectories of surrounding dynamic objects without relying on any static key-point, while also recovering the interaction between surrounding dynamic objects. 
In other words, the movements themselves are the features.

\subsection{3D Multi-Object Tracking (3D MOT)}
3D MOT concerns the detection and tracking of target objects in a video sequence while estimating their 3D locations on the ground~\cite{sharma2018beyond,luiten2020track,osep2017combined}. Most recent works aim to improve tracklet association across frames~\cite{luiten2020track,weng2020gnn3dmot}. 
These approaches, however, assume a simple motion model independent of the camera ego-motion~\cite{weng20203d}, 
which hardly applies to a dynamic observer in a crowd with complex interactions with other pedestrians. 3D MOT in a video with a dynamic observer~\cite{Hu2021QD3DT} has been studied, but the observer motion is known from an external GPS which is often inaccurate~\cite{hery2019}. Our work focuses on reconstructing both the observer ego-motion and surrounding pedestrians in a crowd, while simultaneously learning their complex interactions, which complements these works for visual situational awareness and surveillance.

\section{View Birdification}
Let us first review the task of view birdification~\cite{nishimura2021bmvc}. We have a crowd of people and one observer in the crowd with an ego-centric camera observing the surroundings while moving around. The observer can either be one of the pedestrians of the crowd or a mobile robot, or even an autonomous vehicle, in the crowd. 
As the observer is immersed in the crowd with a limited but dynamic field-of-view, the static background cannot be reliably found in the ego-centric view. 

Let us assume that the crowd consists of $N$ people. We set the z-axis of the world coordinate system to the normal of the ground plane (xy-plane). As in previous work~\cite{nishimura2021bmvc}, we assume the ground plane to be planar and the observer's camera direction is parallel to it. We can assume this without loss of generality as the camera pitch and roll can be corrected either by measurements of the moment (\eg with an IMU) or potentially from optical flow. View birdification thus is the problem of recovering 2D trajectories of the observer and surrounding people (visible in the ego-centric view) on the ground plane (xy-plane) from their 2D in-image movements in the ego-centric view. 

Let $\ped_\pedc = \left [x_\pedc,y_\pedc \right ]^\top$ denote the on-ground location of the $\pedc^{\text{th}}$ pedestrian and $\pose = \left [ c_x, c_y, \theta_z \right ]$ the pose of the observer's camera. The ego-centric camera pose $\pose$ consists of a rotation matrix $R(\theta_z) \in \mathbb R^{2\times 2}$ parameterized by the rotation angle around the z-axis $\theta_z$ and 2D translation $\bm t = -R(\theta_z)\left [c_x, c_y \right ]^\top$, \ie the viewing direction and camera location on the ground, respectively. The observer's camera location is $\left [ c_x, c_y, c_z\right ]^\top$, where the mounted height $c_z$ is constant across frames, and the intrinsic matrix $A \in \mathbb R^{3\times 3}$ is assumed to be constant.

At every timestep $\tau$, we extract the state of each pedestrian $\state^\tau_\pedc$ for all those visible in the observed image, $n \in \{1,2,\dots,N\}$. The pedestrian state encodes the two-dimensional center of the pedestrian's bounding box and the velocity calculated by its displacement from the bounding box center of the previous ($\tau-1 $) frame. These states of visible pedestrians in the ego-centric view $\mathcal S_\pedc^{\tau_1:\tau_2}$ can be extracted with an off-the-shelf multi-object tracker with consistent IDs. Given a sequence of in-image pedestrian states $\States_\pedc^{\tau_1:\tau_2} = \{\state_\pedc^{\tau_1},\state_\pedc^{\tau_1+1},\dots,\state_\pedc^{\tau_2} \}$ from timestep $\tau_1$ to $\tau_2$, our goal is to simultaneously reconstruct the on-ground trajectories of pedestrians $\traj_\pedc^{\tau_1:\tau_2} = \{\ped_\pedc^{\tau_1}, \ped_\pedc^{\tau_1+1},\dots,\ped_\pedc^{\tau_2}\}$ and the observer's camera poses $\Pose^{\tau_1:\tau_2} = \{\pose^{\tau_1},\pose^{\tau_1+1},\dots,\pose^{\tau_2}\}$.

\begin{figure*}[t]
    \centering
    \includegraphics[width=0.7\linewidth]{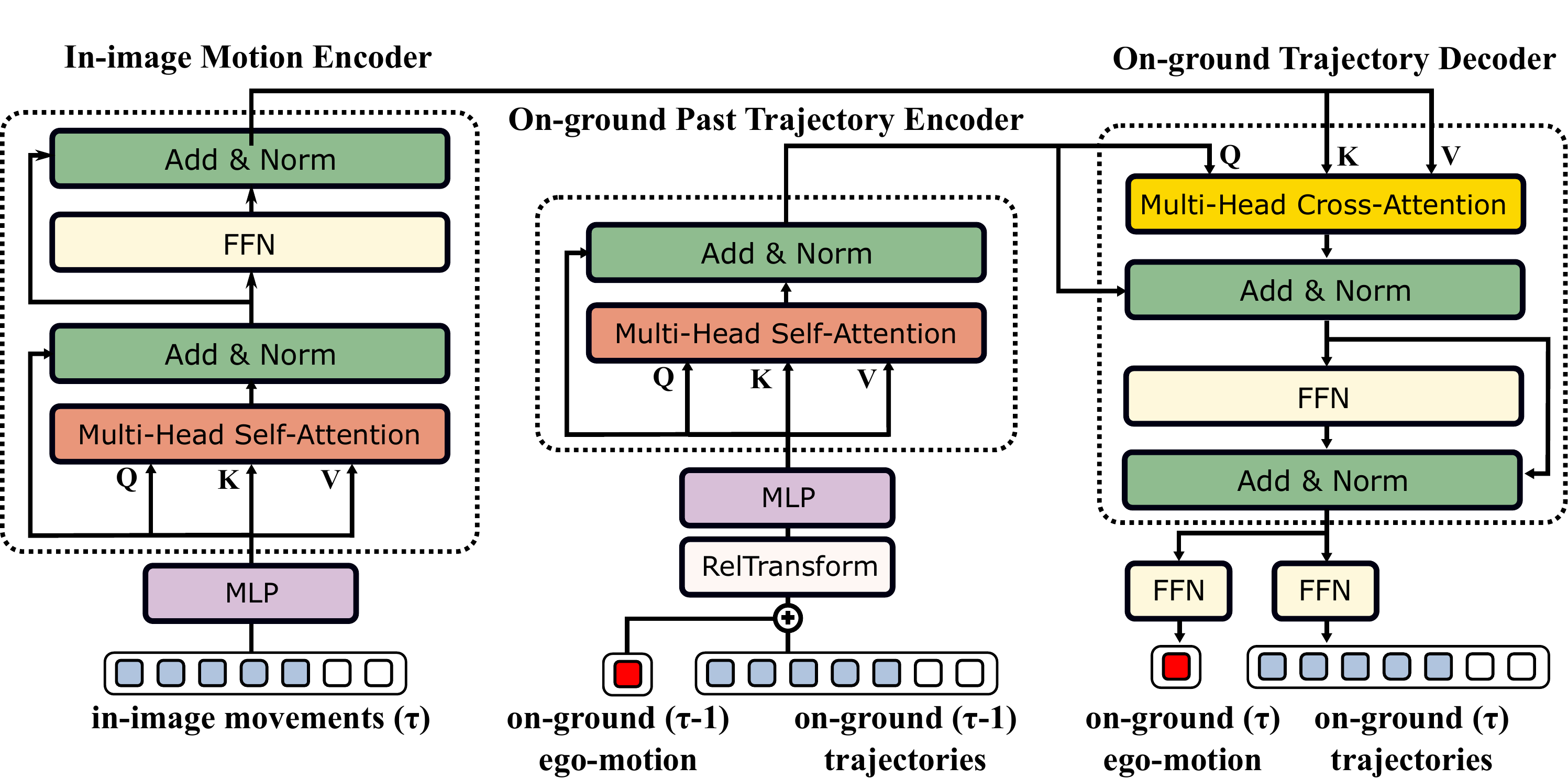}
    \caption{\textbf{The overall architecture of \methodname.} The decoder takes two types of queries: camera queries and pedestrian queries. These queries are fed autoregressively from the previous frame output embeddings of the last decoding layer.}
    \label{fig:architecture}
\end{figure*}

\section{\methodname}
Our goal is to devise a method that jointly transforms the 2D in-image movements into 2D on-ground trajectories and models the on-ground interactions between pedestrians in a single framework. 
For this, we formulate view birdification as a set-to-set translation task, and derive a novel Transformer-based network referred to as \emph{\methodname.} 

\looseness=-1
\subsection{Geometric 2D-to-2D Transformer}\label{sec:geotf}
Given a sequence of in-image pedestrian states for $N$ people in a crowd at time $\tau$, we first embed them into a set of $d$-dimensional state feature vectors $\mathcal F_s \in \mathbb R^{N\times d}$ with a multilayer perceptron (MLP).
We similarly embed past ($\tau-1$) on-ground  trajectories of the pedestrians and the observer's camera, too.
\methodname consists of an encoder that encodes input in-image state features $\mathcal F_s$ into a sequence of hidden state features $\mathcal H_s \in \mathbb R^{N\times d}$, and a decoder that takes in the hidden features and on-ground queries $\mathcal Q_o \in \mathbb R^{(N+1)\times d}$ 
\begin{equation}
    \mathcal H_s = \mathcal E_{\psi} (\mathcal F_s), \quad \mathcal F_o = \mathcal D_{\phi}(\mathcal Q_o,\mathcal H_s)\,,
\end{equation}
where $\mathcal E_\psi$ and $\mathcal D_{\phi}$ are the encoder and decoder models with learnable model parameters $\psi$ and $\phi$, respectively. \Cref{fig:architecture} depicts the overall architecture of our \methodname. 

\paragraph{Attention layers}
A standard attention mechanism~\cite{vaswani2017attention} accepts three types of inputs: a set of queries $\mathcal Q\in \mathbb R^{M\times d}$, a set of key vectors $\mathcal K \in \mathbb R^{N \times d}$, and a set of value embeddings $\mathcal V \in \mathbb R^{N \times d}$.
The output is computed by values weighted by an attention matrix $\bm A \in \mathbb R^{M\times N}$ composed of dot-products of queries and keys, and
we use softmax to normalize the attention weights,
\begin{eqnarray}
  \textrm{Attn}(\mathcal Q, \mathcal K, \mathcal V) = \sum_{j=1}^{N} A_{ij} \bm v_j\,,
  \;
    \bm A_{ij} = \frac{\exp(\bm q_i^\top \bm k_j)}{\sum_{j'=1}^{N}\exp (\bm q_i^\top \bm k_{j'})}\,.
\end{eqnarray}
The query $\bm q$, key $\bm k$, and value $\bm v$ vectors are linear embeddings of the source $\bm f_s$ and target $\bm f_t$ input state features
\begin{equation}
    \bm q = W_q(\bm f_t)\,,\quad \bm k = W_k(\bm f_s)\,, \quad \bm v = W_v(\bm f_s)\,,
\end{equation}
where $W_q$, $W_k$, and $W_v$ are linear embedding matrices specific to the vector types. We refer to the case of $\bm f_s = \bm f_t$ as \emph{self-attention}, and the other case $\bm f_s \neq \bm f_t$ as \emph{cross-attention}.


\paragraph{Attention Mask}
To handle the varying number of pedestrians entering and leaving the observer's view, we apply a mask $\bm M^\tau \in \mathbb R^{M\times N}$ to the attention matrix $\bm A^\tau$ as $M^\tau \odot \bm A^\tau$, where $\odot$ denotes Hadamard product.
The element of the mask $M_{ij}^\tau$ is set to $\boldsymbol{0}$ if either of the pedestrians $i$ or $j$ are missing at time $\tau$, otherwise $\boldsymbol{1}$. This allows us to handle temporarily occluded pedestrians. For more details, please refer to Sec. 3 of the supplementary material.

\paragraph{In-Image Motion Encoder}
The encoder architecture consists of a single multi-head self-attention layer~\cite{vaswani2017attention} and a feed-forward network (FFN) layer. 
We define the input pedestrian states as $\state_i^\tau = [p_x,p_y,\Delta p_x, \Delta p_y]^\top$, consisting of the 2D center of the detected bounding box $\bm p = [p_x,p_y]^\top$ and is its velocity $\Delta \bm p = [\Delta p_x,\Delta p_y]^\top$.
The encoder $\mathcal E_\psi$ computes self-attention over all queries generated by input state feature embeddings $\mathcal F_s$, which encodes the interactions between observed pedestrians in image space.

\paragraph{On-ground Trajectory Decoder}
The Transformer decoder $\mathcal D_\phi$ integrates the self-attention based on-ground motion model and the cross-attention between on-ground and ego-views. 
First, the \textbf{On-Ground Past Trajectory Encoder} applies self-attention over queries $\mathcal Q$ consisting of an ego-motion query $\bm q_{\pi}^{\tau-1}$ and on-ground pedestrian trajectory queries $\{\bm q_1^{\tau-1},\dots,\bm q_N^{\tau-1}\}$ extracted from previous estimates at $\tau-1$. We calculate these with on-ground queries $\bm q_{\pi}^{\tau-1}= W_q(\mathrm{MLP}(\Delta \bm \pi^{\tau-1}))$ and $\bm q_{n}^{\tau-1} = W_q(\mathrm{MLP}(\ped_\pedc^{\tau-1} \oplus \Delta \ped_\pedc^{\tau-1}))$, respectively. The attention learns to capture the implicit local and global interactions of all pedestrians to better predict the future location from past trajectories. 
Second, the cross-attention layer accepts hidden state features $\mathcal H_s$ processed by the encoder and on-ground trajectory queries $\mathcal Q_o$ processed by the self-attention layer. This layer outputs feature embeddings $\mathcal F_o \in \mathbb R^{(N+1)\times d}$ by incorporating features from the ego-centric view. The output $\mathcal F_o$ is decoded to the camera ego-motion $\Pi^\tau$ and $N$ pedestrian trajectories $\{\traj_1^{\tau},\dots, \traj_N^{\tau}\}$ by task-specific heads. The trajectory decoder is autoregressive, which outputs trajectory estimates one step at a time and feeds the current estimates back into the model as queries to produce the trajectories of the next timestep.

\subsection{Relative Position Transformation}\label{sec:reltransform}
A key challenge of view birdification lies in the inconsistency of coordinate systems between input and output trajectories. Unlike conventional frame-by-frame 2D-to-3D lifting~\cite{bertoni2019monoloco} or image-based bird's eye-view transformation~\cite{yang2021projecting}, once the viewpoint of the observer's camera is changed, the observed movements of pedestrians in the image change dramatically. To encourage the network to generalize over diverse combinations of trajectories and observer positions, we transform all the on-ground pedestrian queries relative to the previous $\tau-1$ observer's camera estimates at every timestep $\tau$,
\begin{eqnarray}
  \tilde {\ped}_\pedc^{\tau} &=& R(\theta_z^{\tau-1})\ped_\pedc^{\tau} + \bm t^{\tau-1}\,,
  \\
  \Delta \tilde \ped_\pedc^\tau &=& R(\theta_z^{\tau-1})(\ped_\pedc^{\tau}-\ped_\pedc^{\tau-1})\,,
\end{eqnarray}
where $\bm t^{\tau-1} = -R(\theta_z^{\tau-1}) [c_x^{\tau-1}, c_y^{\tau-1}]^\top$ is the camera translation. We force all on-ground trajectory coordinates to be centered on the observer's camera by defining positions and velocities relative to the observer's camera $\tilde \ped \oplus {\Delta \tilde \ped}$ as pedestrian features, and the camera displacements $\Delta \bm \pi = [\Delta c_x, \Delta c_y, \Delta \theta_z ]^\top$ as the observer's feature.

\subsection{Ego-motion Estimation by Task-specific Heads}\label{sec:ego}
To achieve simultaneous recovery of pedestrian trajectories and ego-motion of the observer's camera, we formulate birdification as a multi-task learning problem. Given a set of past queries $\{\bm q_c^{\tau-1}, \bm q_1^{\tau-1},\dots, \bm q_N^{\tau}\}$ consisting of trajectories of the observer and surrounding pedestrians, the decoder transforms the joint set of camera and pedestrian queries into output embeddings $\mathcal F_o \in \mathbb R^{(N+1)\times d}$. The output embeddings $\mathcal F_o$ consist of two types of features: (i) ego-motion embedding $\mathcal F_{\mathrm{ego}} \in \mathbb R^{1 \times d}$ from which the motion of the observer's camera on the ground is recovered, and (ii) pedestrian trajectory embeddings $\mathcal F_{\mathrm{traj}} \in \mathbb R^{N \times d}$ represented in a relative coordinate system, where the origin is the position of the camera. These two queries calculated from the previous $t-1$ frame are decoded simultaneously. We define individual loss functions for these two tasks.

\paragraph{Ego-Motion Loss}
The ego-motion output embedding $\mathcal F_{\mathrm{ego}}$ is decoded into $\Delta {\bm \pi} = [\Delta c_x, \Delta c_y, \Delta \theta_z]^\top$ by a single feed-forward network. For a batch $\{ \Delta \pose^{\tau}, \dots, \Delta \pose^{T} \} \in \mathbb R^{3\times T}$ of duration $T$,
we compute the mean squared error
\begin{equation}
        \mathcal L_{\mathrm{ego}} = \sum_{\tau=1}^{T} \| \Delta \dot{\pose}^{\tau}  - \Delta {\pose}^{\tau} \|\,,
\end{equation}
where $\dot{\pose}$ is the ground-truth camera pose of an observer.

\paragraph{Pedestrian Trajectory Loss}
Pedestrian trajectory embeddings $\mathcal F_{\mathrm{traj}}$ are decoded into 2D positions and velocities $\tilde \ped \oplus \Delta \tilde \ped \in \mathbb R^{4}$ relative to the observer's camera. Given a batch of $N$ observed pedestrians for duration $T$, we define the trajectory loss function as
\begin{equation}
  \begin{split}
    \mathcal L_{\mathrm{traj}} = \sum_{\tau=1}^{T} \sum_{\pedc=1}^{N} &\| \dot \ped_\pedc^{\tau} - (R(\theta_z^{\tau-1})^\top (\tilde \ped_\pedc^{\tau} - \bm t^{\tau-1}) \| \\
    + &\| \Delta \dot \ped_\pedc^{\tau} - R(\theta_z^{\tau-1})^\top \Delta \tilde \ped_\pedc^{\tau} \|\,,
  \end{split}
\end{equation}
where the output estimate $\ped_i^\tau$ is transformed into the world coordinate system by the camera pose estimates consisting of the rotation angle $\theta_z^\tau =\theta_z^{\tau-1} + \Delta \theta_z^\tau$ and 2D translation $\bm t^{\tau} = R(\Delta \theta_z^\tau)\bm t^{\tau-1} + \Delta \bm t^\tau$.

\looseness=-1
\paragraph{Observer Reprojection Loss}
What makes view birdification unique from other on-ground trajectory modeling problems is its ego-centric view input. Although the 2D ego-centric view degenerates depth information of the observed pedestrian movements, it also provides a powerful inductive bias for on-ground trajectory estimates. 
Using the oberver's camera intrinsic matrix $A$, we compute the reprojection loss in the image plane
\begin{equation}
      \mathcal L_{\mathrm{proj}} = \sum_{\tau=1}^{T}\sum_{\pedc=1}^{N} \left \| \overline{ \pedi}_\pedc^\tau - sA \overline{\ped}_i^\tau  \right \|\,,
\end{equation}
where $\overline{\pedi} = \left [ p_x, p_y, 1 \right ]^\top$ is the homogeneous coordinate of the observed 2D bounding box center, and $\overline{\ped} = \left [x, y, h/2 \right ]^\top$ is the half point of the pedestrian height standing on the position $\ped_\pedc = R(\Delta \theta_z) \tilde{\ped}_\pedc + \Delta \bm t$, respectively. The scaling factor $s$ is determined by normalizing the $z-$value of the projected point in the image.

\paragraph{Total Loss}
The complete multi-task loss becomes
\begin{equation}
  \label{eq:total_loss}
    \mathcal L = \mathcal L_{\mathrm{traj}} + \lambda_1 \mathcal L_{\mathrm{ego}} + \lambda_2 \mathcal L_{\mathrm{proj}}\,.
\end{equation}
To facilitate stable training, we apply curriculum learning to the reprojection loss weight $\lambda_2$. We set $\lambda_2 = 0$ for the first $200$ epochs, and switch to $\lambda_2 > 0$ for the rest of the epochs.

\paragraph{Test-time refinement}
The reprojection loss can be used to refine the ego-motion towards the pedestrian trajectory estimates at inference time. That is, we incorporate the reprojection errors into our network as a soft geometric constraint \ie weighted reprojection loss, in the training phase, and as a hard geometric constraint at inference time.

\section{Experiments}

\begin{table*}[t]
\vspace{12pt}
  \centering
  \small
  \begin{tabularx}{0.85\linewidth}{lllllllll}
    \toprule[1pt]
     &&\multicolumn{2}{c}{\textbf{Hotel / sparse}} & \multicolumn{2}{c}{\textbf{ETH / mid}} & \multicolumn{2}{c}{\textbf{Students / dense}} \\
   && $\XerrRel$ [m]&  $\Xerr$ [m]&$\XerrRel$ [m]&  $\Xerr$ [m]& $\XerrRel$ [m]&  $\Xerr$ [m]\\
    \toprule[1pt]
TransMotion-I & & -- & 0.183  & -- & 0.201 & -- & 0.216  \\
TransMotion-C  && -- & 0.106 & -- & 0.223 & -- & 0.211 \\
    \midrule
\baselinename-CV~\cite{nishimura2021bmvc} && \textbf{0.051}& 0.070 & 0.089 & 0.115 & 0.023 & 0.024 \\
        \baselinename-SF~\cite{nishimura2021bmvc} && \textbf{0.048}$^*$ & \textbf{0.052}$^*$ & \textbf{0.070}$^*$  & \textbf{0.079}$^*$ & \textbf{0.009}$^*$  & \textbf{0.010}$^*$\\
    \midrule
\textbf{ViewBirdiformer-I}& & 0.123 & 0.123 & 0.170 & 0.170 & 0.071 & 0.071 \\
\textbf{ViewBirdiformer-C} & & 0.097 & 0.098 & 0.216 & 0.217 & 0.058 & 0.059 \\
   \midrule
\rowcolor[rgb]{0.93,1.0,0.87} \textbf{ViewBirdiformer-I} & w/post-processing & 0.062 & 0.081 & 0.087 & 0.102 & \textbf{0.010} & \textbf{0.010}$^*$ \\
\rowcolor[rgb]{0.93,1.0,0.87} \textbf{ViewBirdiformer-C} &w/post-processing & 0.071 & 0.092 & 0.099 & 0.115 & \textbf{0.010} & \textbf{0.011}  \\
    \midrule[1pt]\midrule[1pt]
         && $\Rerr$ [rad]&  $\Terr$ [m]&$\Rerr$ [rad]& $\Terr$ [m] &$\Rerr$ [rad]& $\Terr$[m]\\
    \toprule[1pt]
    \baselinename-CV~\cite{nishimura2021bmvc}& & \textbf{0.015} & 0.066 & \textbf{0.016} & 0.095 & \textbf{0.001}$^*$ & \textbf{0.010} \\
        \baselinename-SF~\cite{nishimura2021bmvc}& &  \textbf{0.015} & \textbf{0.062} & \textbf{0.015}$^*$ & \textbf{0.089} & \textbf{0.001}$^*$ & \textbf{0.009}$^*$ \\
    \midrule
\textbf{ViewBirdiformer-I}  & & 0.125 & 0.085 & 0.032 & 0.093 & 0.061 & 0.068 \\
\textbf{ViewBirdiformer-C} && 0.063 & 0.091 & 0.101 & 0.098 & 0.080 & 0.069 \\
    \midrule
\rowcolor[rgb]{0.93,1.0,0.87}
   \textbf{ViewBirdiformer-I} & w/post-processing & \textbf{0.014}$^*$ & \textbf{0.059}$^*$ & \textbf{0.015}$^*$ & \textbf{0.091} & \textbf{0.002} & \textbf{0.011} \\
 \rowcolor[rgb]{0.93,1.0,0.87} \textbf{ViewBirdiformer-C}& w/post-processing & \textbf{0.016} & \textbf{0.061} & 0.021 & 0.098 & \textbf{0.002} & \textbf{0.011} \\
     \bottomrule
  \end{tabularx}
  \caption{\textbf{Quantitative Results.} The top table shows relative and absolute localization errors of pedestrian trajectories, $\Delta \tilde x$ and $\Delta x$. The motion model baseline only extrapolates the on-ground movement and thus results in missing entries (-) in $\Delta \tilde x$. The bottom table shows the camera ego-motion errors $\Delta r$ and $\Delta \bm t$. We highlight the best ($^*$) and similar to best (accuracy gap $\leq 0.005$) results of localization accuracy. The results demonstrate the effectiveness of our proposed \methodname.}
    \label{tab:view-birdification-benchmark}
\end{table*}

\subsection{View Birdification Datasets}
We evaluate our method on view birdification data consisting of paired real pedestrian trajectories and synthetic ego-views of them.
The dataset is generated from public pedestrian trajectory datasets ETH~\cite{pellegrini2009you} and UCY~\cite{lerner2007crowds} by following the instructions of the original view birdification paper~\cite{nishimura2021bmvc}. To generate a sufficient amount of ego-views including diverse patterns of projected movements, we mount a virtual, perspective camera on each of the pedestrians (\ie an observer) in turn. As a result, we obtain paired trajectories and their ego-views for as many as the number of pedestrians in each scene. Following previous work~\cite{nishimura2021bmvc}, we assume ideal observation, \ie pedestrians are not occluded by each other and projected heights can be deduced from the observed images.
There are three datasets named after the scenes they capture, \textbf{Hotel}, \textbf{ETH}, and \textbf{Students}, which correspond to sparse, moderate, and dense crowds, respectively. We prepare two types of splits of the view birdification dataset. The first one is (i) intra-scene validation split. For each scene, train, val, and test splits are generated. This allows evaluation of how \methodname generalizes to unseen trajectories. The second one is (ii) cross-scene validation split. We pick one scene for testing and choose the rest of the remaining scenes for validation and training. These splits allow evaluation of how \methodname generalizes to unknown scenes.

\paragraph{Evaluation Metric}
Our proposed framework first reconstructs the ego-motion of the observer and the trajectories of her surrounding pedestrians in the observer's camera coordinate system. 
The absolute positions and trajectories of the pedestrians in the world coordinate system are computed by coupling these two outputs, \ie $\ped_{\pi}^\tau = R(\theta_z^{\tau-1} + \Delta \theta_z^\tau)\tilde \ped_i^\tau + R(\Delta \theta_z^\tau) \bm t^{\tau-1} + \Delta \bm t^{\tau}$. 
We evaluate the accuracy of our method by measuring the differences of the estimated positions of pedestrians $\ped$ and the ego-motion of the observer $\Delta \Pi = (\Delta \bm t, \Delta \theta_z)$ from their corresponding ground truths $\dot \ped$, $\Delta \dot{ \bm t}$, and $\Delta \dot \theta_z$. The translation error of the observer is $\Terr = \frac{1}{T}\sum \|\ped_{\pi}^\tau - \dot{\ped}_{\pi}^\tau\|$, where $T$ denotes the duration of a sequence. The rotation error of the observer is $\Rerr = \frac{1}{T} \sum_{\tau} \arccos (\frac{\text{tr}\left(R (\Delta \dot  \theta_z^{\tau})R(\Delta \theta_z^{\tau})^\top\right)-1}{2})$, where $\text{tr}$ is the matrix trace. We also evaluate the absolute and relative reconstruction errors of pedestrians by $\Delta \ped = \frac{1}{N}\frac{1}{T}\sum_{i}\sum_{\tau}\|\ped_i^\tau -\Xgt_i^\tau\| $ and $\Delta \tilde \ped = \frac{1}{N}\frac{1}{T} \sum_\pedc \sum_{\tau} \| \tilde \ped_i^\tau - R(\theta_z^{\tau-1})\Xgt_\pedc - \bm t^{\tau-1} \|$ .

\paragraph{Baseline Methods}
We compare our method with a purely geometric view birdification approach~\cite{nishimura2021bmvc}, the only other view birdification method. 
We use the parameter values from the original paper, which we refer to as \emph{\baselinename-CV} and \emph{\baselinename-SF} based on the assumed motion model: Constant Velocity (CV)~\cite{scholler2020constant} and Social Force (SF)~\cite{helbing1995social}, respectively. We also evaluate the effectiveness of the ego-view encoder and the cross-attention by comparing with the direct use of an on-ground motion model which takes $\tau-1$ on-ground trajectories as inputs and simply predicts positions and velocities $\ped^\tau \oplus \Delta \ped^\tau$ for $\tau$. For this, we train a simple Transformer-based motion model with one multi-head self-attention layer which we refer to as \emph{\motionmodelname}. Note that, although \emph{\methodname} and \emph{\baselinename} both take as inputs the ego-centric view at time $\tau$ and the past on-ground trajectory estimates at time $\tau-1$, \emph{\motionmodelname} only takes past on-ground trajectory estimates.

We consider two variants of \methodname. The first, \emph{ViewBirdiformer-I}, is trained on the intra-scene validation split, and the second, \emph{ViewBirdiformer-C}, on the cross-scene validation split. Similarly, simple motion models composed of single-layer self-attention Transformers each trained with these validation splits are referred to as \emph{TransMotion-I} and \emph{TransMotion-C}, respectively.

\begin{figure*}[t]
    \centering
    \includegraphics[width=0.9\linewidth]{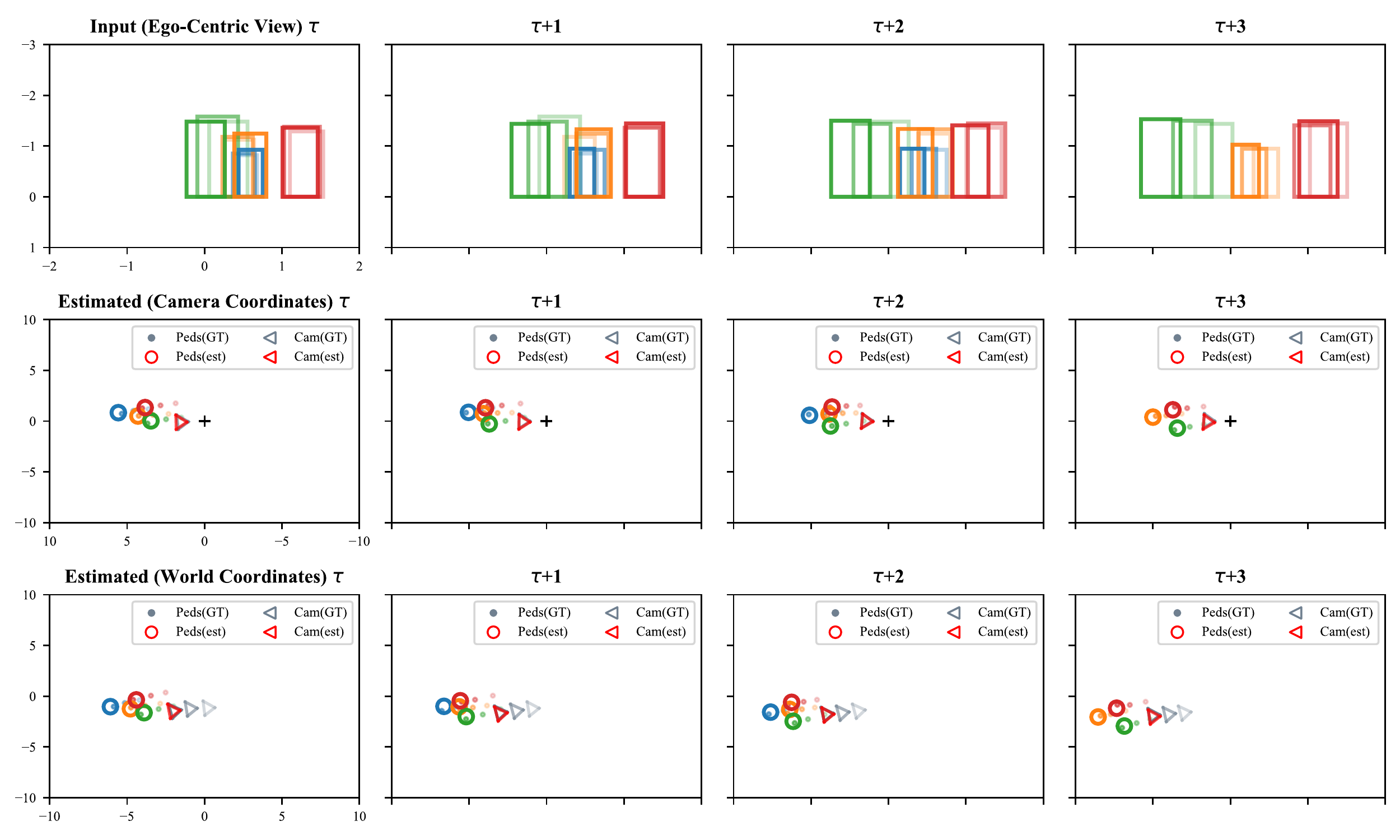}
    \caption{\textbf{Qualitative Results of \methodname-I without post-processing applied to ETH datasets.} The top row shows the input bounding boxes, where the same color box corresponds to the same pedestrian ID and the boxes with low alpha values correspond to the past $\tau-1$ frame positions. The second row shows the reconstructed camera pose and pedestrian locations at time $\tau$ in the $\tau-1$ camera-centric coordinates. ``+" depicts the origin of the camera coordinate system. These relative observations are converted to the world coordinates by the estimated camera pose at every frame (the third row). Grey triangles and circles denote ground-truth camera and pedestrian positions, respectively. 
    These results show that our method successfully birdifies input bounding box movements into on-ground trajectories very accurately. More results are provided in the supplementary material.
    }
    \label{fig:qualitative_results}
\end{figure*}

\begin{table*}[t]
\vspace{6pt}
    \centering
    \small
    \renewcommand{\arraystretch}{0.85}
    \begin{tabularx}{0.95\linewidth}{lccccccccc}
    \toprule[1pt]
       \multirow{2}{*}{Dataset} &\multicolumn{3}{c}{\textbf{Hotel / sparse}} & \multicolumn{3}{c}{\textbf{ETH / mid}} & \multicolumn{3}{c}{\textbf{Students / dense}} \\
       & $\XerrRel$ [m]&  $\Rerr $ [rad]&$\Terr $ [m]&  $\XerrRel$ [m]&  $\Rerr $ [rad]&$\Terr $ [m] & $\XerrRel$ [m]&  $\Rerr $ [rad]&$\Terr $ [m] \\
    \toprule[1pt]
       w/o RelTransform & 2.115& 0.055 & 0.277 & 2.105& 0.053 & 0.279 & 1.713& 0.090 & 0.269 \\
       w/o ReprojectionLoss & 0.148 & 0.148 & 0.197 & 0.180 & 0.038 & 0.179 & 0.081 & 0.065 & 0.111\\
       \midrule
 \rowcolor[rgb]{0.93,1.0,0.87}       \textbf{Ours}  & \textbf{0.123} & \textbf{0.125} & \textbf{0.085} & \textbf{0.170} & \textbf{0.032} & \textbf{0.093} & \textbf{0.071} & \textbf{0.061} & \textbf{0.068} \\ 
     \bottomrule
    \end{tabularx}
    \caption{\textbf{Ablation Studies.} w/o denotes our proposed architecture without the specified component. The results demonstrate that relative transformation of the decoder inputs (\Cref{sec:reltransform}) is essential for accurate localization of surrounding pedestrians, and the additional reprojection loss is key to accurate ego-motion estimation.} 
    \label{tab:ablation_study}
\end{table*}

\paragraph{Implementation Details}
All networks were implemented in PyTorch. The camera intrinsic matrix $A$ was set to that of a generic camera with FOV=$120^\circ$ and $f=2.46$. Both the embedded dimension of the on-ground trajectories and in-image movements, $d$ is set to $32$. We use an MLP with $16$ hidden units for embedding input features. The number of heads for the multi-head attention layer is all set to $8$. Loss coefficient $\lambda_1$ is set to 1.0 and $\lambda_2$ is set to 0.3 after $200$ epochs. We use Adam optimizer and set the constant learning rate to $0.001$ for all epochs. All the models are trained with a single NVIDIA Tesla V100 GPU and Intel Xeon Gold 6252 CPU. The training time is approximately $3$ hours for the train split excluding Students and $14$ hours for that including Students. For all the datasets, we transformed trajectories into scene-centered coordinates so that the origin of the mean position of all the pedestrians is $0$. 
The outputs of our proposed network are post-processed by the test-time refinement described in \cref{sec:ego}.

\subsection{Comparison with Geometric Baseline}
\looseness=-1
\paragraph{Localization Accuracy}
\Cref{tab:view-birdification-benchmark} shows quantitative results. GeoVB~\cite{nishimura2021bmvc} achieves high accuracy by iteratively optimizing the camera ego-motion and pedestrian positions by densely sampling possible positions for every frame. Although the accuracy of our \methodname is slightly lower, it achieves sufficiently high absolute accuracy (\eg 5cm errors in $20 \times 20$ m field) with a single inference pass. \Cref{fig:qualitative_results} visualizes qualitative results of our method on a typical crowd sequence, which clearly shows that our method reconstructs accurate on-ground trajectories. Even with the cross-scene validation split, \emph{ViewBirdiformer-C} achieves comparable results. By incorporating the geometric refinement at inference time, \methodname achieves comparable or superior accuracy to the state-of-the-art~\cite{nishimura2021bmvc} but still in three orders of magnitude shorter time.

\begin{figure}[t]
    \centering
    \includegraphics[width=0.9\linewidth]{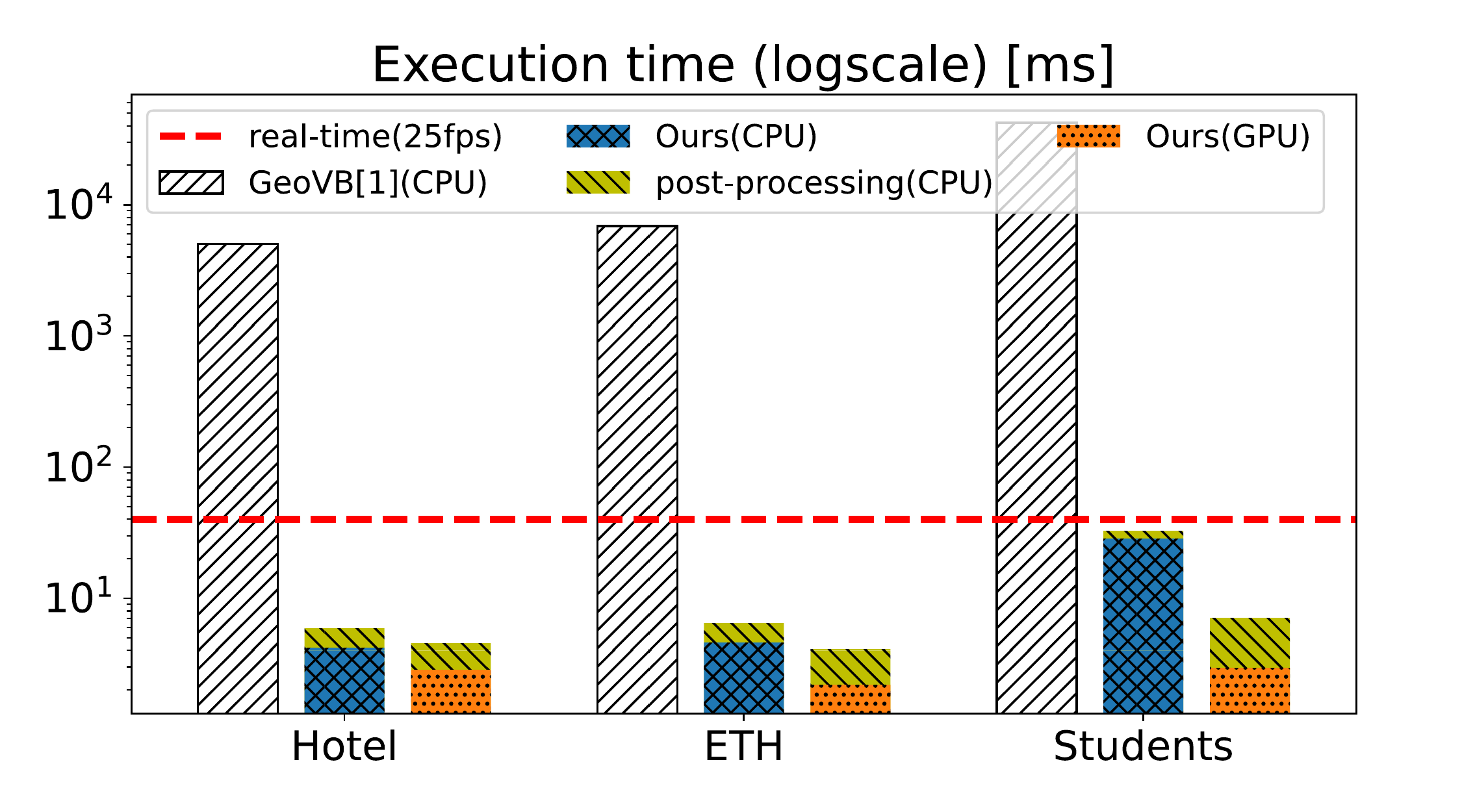}
    \caption{\textbf{Execution time.} We measure the execution times of our method on a CPU and a GPU. The post-processing is executed on the CPU. These results are averaged over the samples of each dataset. Our proposed approach achieves three orders of magnitude reduction in execution time for the same accuracy in comparison to past approach~\cite{nishimura2021bmvc} even including post-processing.}
    \label{fig:execution_time}
\end{figure}

\paragraph{Efficiency of \methodname}
\Cref{fig:execution_time} shows the execution time of our method and GeoVB~\cite{nishimura2021bmvc} on a single Intel Core i5-7500 CPU and a NVIDIA GeForce 1080Ti GPU. These results clearly demonstrate the efficiency of our method compared to GeoVB. The unified transformer architecture of our \methodname enables estimation of both ego-motion and pedestrian trajectories with a single inference pass without the costly iteration process in GeoVB~\cite{nishimura2021bmvc}, which results in three orders of magnitude improvement in execution time. 
For $N$ pedestrians, $S$ samples, and $T$ iterations, the computational complexity of GeoVB is $\mathcal O(NS^2T)$ and it is hardly parallelizable as it requires sequential update over all possible samples $S$ ($S \gg N$).
In contrast, the computational complexity of \methodname is $\mathcal O(N^2d)$~\cite{vaswani2017attention} and its implementation can naturally be parallelized within a GPU, which collectively realize this significant reduction in execution time. 
Most important, even with the geometric refinement at inference time, \methodname achieves accuracy on par with the state-of-the-art~\cite{nishimura2021bmvc} while maintaining this orders of magnitude faster execution.

\subsection{Ablation Studies}

\paragraph{Cross-Attention Between Views}
\Cref{tab:view-birdification-benchmark} compares the accuracy of \methodname and simple extrapolation of on-ground movements using dedicated simple transformers. While \methodname takes the current ego-centric view and the past on-ground trajectory estimates as inputs, \motionmodelname only takes the past trajectory estimates as inputs. \methodname shows superior performance over \motionmodelname in pedestrian localization. These results clearly show that the cross-attention mechanism between on-ground motions and movements in the ego-centric views is essential for accurate trajectory estimation of the surrounding pedestrians.

\paragraph{Relative Position Transformation}
\Cref{tab:ablation_study} shows the results of ablating the relative position transforms (\Cref{sec:reltransform}). All models are trained with the intra-scene split of the birdification dataset to avoid generalization errors of the learnt motion model. \emph{w/o RelTransform} takes on-ground trajectories in world coordinates as decoder inputs. Without the relative position transformations described in \cref{sec:geotf}, the proposed framework shows significant accuracy drops, especially in pedestrian localization. This is likely caused by the inconsistency of the coordinate system between on-ground past trajectory inputs and egocentric view inputs and demonstrates the importance of the relative transformation for generalization of the model. 

\paragraph{Reprojection Loss}
\emph{w/o ReprojectionLoss} in \cref{tab:ablation_study} considers only the ego-motion loss and the pedestrian trajectory loss, \ie $\lambda_2=0$ in \cref{eq:total_loss}. The results show that the reprojection loss slightly improves the accuracy of ego-motion estimates. This is because the reprojection loss works similarly to geometric constraints as in \emph{\baselinename}.

\subsection{Limitations and Degenerate Scenario}
If the observed relative movements are static (\ie an observer is following the pedestrian at the same speed), our model cannot break the fundamental ambiguity. Such degenerate scenarios, however, rarely happen in crowds as there will be other pedestrians. Our method also assumes that the heights of pedestrians are more or less the same and that the detected bounding boxes are correct. We plan to relax these requirements by developing an end-to-end framework that handles both tracking and birdification on the ground plane from the raw image inputs in our future work.

\section{Conclusion}
In this paper, we introduced \methodname for view birdification. 
The proposed architecture enables efficient and accurate view birdification by adaptively attending to movement features of the observer and pedestrians in the image plane and on the ground. Extensive evaluations demonstrate the effectiveness of \methodname for crowds with diverse pedestrian interactions. 
We believe \methodname finds use in various applications of crowd modeling and synthesis across a wide range of disciplines. We plan to release our code and data to catalyze such use.

\balance

\bibliographystyle{IEEEtran}
\bibliography{egbib}

\end{document}